\documentclass{article}

\usepackage{arxiv}

\usepackage[utf8]{inputenc} 
\usepackage[T1]{fontenc}    
\usepackage{hyperref}       
\usepackage{url}            
\usepackage{booktabs}       
\usepackage{amsfonts}       
\usepackage{nicefrac}       
\usepackage{microtype}      
\usepackage{lipsum}
\usepackage{boldline}
\usepackage[export]{adjustbox}
\usepackage[caption=false]{subfig}
\usepackage{adjustbox}
\usepackage{colortbl}
\usepackage{float}
\usepackage{wrapfig}
\usepackage{lipsum}
\usepackage{stackengine}
\usepackage{graphicx}
\usepackage{siunitx}
\usepackage{algorithm}
\usepackage{algpseudocode}%
\usepackage{amsmath}
\title{Adaptive Neuro-Surrogate-Based Optimisation Method for Wave Energy Converters Placement Optimisation}

\author{
  Mehdi Neshat \\
  Optimization and Logistics Group\\
  School of Computer Science\\
  The University of Adelaide\\
   Australia \\
  \texttt{mehdi.neshat@adelaide.edu.au} \\
  \And
   Ehsan Abbasnejad \\
   The Australian Institute for Machine Learning\\
   University of Adelaide \\
   Australia \\
   \texttt{ehsan.abbasnejad@adelaide.edu.au} \\
   \And
    Qinfeng Shi \\
   The Australian Institute for Machine Learning\\
   University of Adelaide \\
   Australia \\
   \texttt{qinfeng.shi@adelaide.edu.au} \\
   \And
  Bradley Alexander \\
  Optimization and Logistics Group\\
  School of Computer Science\\
  The University of Adelaide\\
   Australia \\
  \texttt{bradley.alexander@adelaide.edu.au} \\
   \And
  Markus Wagner\\
  Optimization and Logistics Group\\
  School of Computer Science\\
  The University of Adelaide\\
   Australia \\
  \texttt{markus.wagner@adelaide.edu.au} \\
}

\begin{document}
\maketitle

\begin{abstract}
The installed amount of renewable energy has expanded massively in recent years. Wave energy, with its high capacity factors has great potential to complement established sources of solar and wind energy.
This study explores the problem of optimising the layout of advanced, three-tether wave energy converters in a size-constrained farm in a numerically modelled ocean environment. 
Simulating and computing the complicated hydrodynamic interactions in wave farms can be computationally costly, which limits optimisation methods to have just a few thousand evaluations. For dealing with this expensive optimisation problem, an adaptive neuro-surrogate optimisation (ANSO) method is proposed that consists of a surrogate Recurrent Neural Network (RNN) model trained with a very limited number of observations. This model is coupled with a fast meta-heuristic optimiser for adjusting the model's hyper-parameters.  The trained model is applied using a greedy local search with a backtracking optimisation strategy. For evaluating the performance of the proposed approach, some of the more popular and successful Evolutionary Algorithms (EAs) are compared in four real wave scenarios (Sydney, Perth, Adelaide and Tasmania). Experimental results show that the adaptive neuro model is competitive with other  optimisation methods in terms of total harnessed power output and faster in terms of total computational costs.
\end{abstract}


\section{Introduction}
As the global demand for energy continues to grow, the advancement and deployment of new green energy sources are of paramount significance. Due to high capacity factors and energy densities compared to other renewable energy sources, ocean waves energy has attracted research and industry interest for a number of years~\cite{drew2009review}.
Wave Energy Converters (WEC's) are typically laid out in arrays and, to maximise power absorption, it is important to arrange them carefully with respect to each other \cite{de2014factors}.
The number of hydrodynamic interactions increases quadratically  with the number of WEC's in the array. Modelling these interactions for a single moderately-sized farm layout can take several minutes. Moreover, the optimisation problem for farm-layouts is multi-modal--typically requiring the use of many evaluations to adequately explore the search space. 
There is scope to improve the efficiency of the search process through the use of a learned surrogate model. The challenge is to train such a model fast enough to allow an overall reduction in optimisation time.
This paper proposes a new hybrid adaptive neuro-surrogate model (ANSO) for maximizing the total absorbed power of WECs layouts in detailed models of  four real wave regimes from the southern coast of Australia (Sydney, Adelaide, Perth and Tasmania). 
Our approach utilises a neural network that acts as a surrogate for estimating the best position for placement of the converters. 
The key contributions of this paper are: 
\begin{enumerate}
\item Designing a neuro-surrogate model for predicting total wave farm energy by training of recurrent neural network (RNNs) using data accumulated from evaluations of farm layouts.
\item The use of the Grey Wolf Optimiser~\cite{mirjalili2020grey} to continuously tune hyper-parameters for each surrogate. 
\item  A new symmetric local search heuristic with greedy WEC position selection combined with a backtracking modification (BO) to improve the layouts further for delicate adjustments.
\end{enumerate}
We demonstrate that the adaptive framework described outperforms previously published results in terms of both optimisation speed (even when total training time is included) and total absorbed power output for 16-WEC layouts. 
\subsection{Related work}
In this application domain, neural networks have been utilized for predicting the wave features (height, period and direction) more than other ML techniques~\cite{bhattacharya2003neural}. In early work, Alexandre et al.~\cite{alexandre2015hybrid} applied a hybrid Genetic Algorithm (GA) and an extreme learning machine (ELM) (GA-ELM) for reconstructing missing parameters from readings from nearby sensor buoys. The same study~\cite{cornejo2016grouping} investigated a combination of the grouping GA and ELM (GGA-ELM) for feature extraction and wave parameter estimation. 
A later approach~\cite{cornejo2018bayesian}, combined the GGA--ELM with Bayesian Optimisation (BO) for predicting the ocean wave features. BO improved the model significantly at the cost of increased computation time.  
Sarkar et al.~\cite{sarkar2016prediction} combined machine learning and optimisation of arrays of, relatively simple, oscillating surge WECs. They were able to use this technique to effectively optimise arrays of up to 40 WEC's -- subject to fixed spacing constraints. 
Recently, James et al.~\cite{james2018machine} used two different supervised ML methods (MLP and SVM) to estimate WEC layout performance and characterise the wave environment~\cite{james2018machine}. However, the models
produced required a large training data-set and manual tuning of hyper-parameters. 

In work optimising WEC control parameters, Li et al.~\cite{li2018maximization} trained a feed-forward neural network (FFNN) to learn key temporal relationships between wave forces. While the model required many samples to train it exhibited high accuracy and was used effectively in parameter optimisation for the WEC controller.
Recently, Lu et al.~\cite{lu2019recurrent} proposed a hybrid WECs PTO controller which consists of a recurrent wavelet-based Elman neural network (RWENN) with an online back-propagation training method and a modified gravitational search algorithm (MGSA) for tuning the learning rate and improving learning capability. The method was used to control the rotor acceleration of the combined offshore wind and wave power converter arrangements.
Finally, recent work by Neshat et al.~\cite{neshat2018detailed} evaluated a wide variety of EAs and hybrid methods by utilizing an irregular wave model with seven wave directions and found that a mixture of a local search combined with the Nelder-Mead simplex method achieved the best array configurations in terms of the total power output.

\section{Wave Energy Converter Model} \label{WECS-Model}
We use a WEC hydrodynamic model for a fully submerged three-tether buoy. Each tether is attached to a converter installed on the seafloor ~\cite{scruggs2013optimal}. The relevant details of the WECs modelled in this research are: Buoy number=$16$, Buoy radius=\SI{5}{\metre}, Submergence depth=\SI{3}{\metre}, Water depth=\SI{30}{\metre}, Buoy mass=\SI{376}{\tonne}, Buoy volume=\SI{523.60}{\metre\squared}  and Tether angle=\ang{55}.

\subsection{System dynamics and parameters}
\begin{figure}[t]
\centering
\subfloat[]{
\includegraphics[clip,width=0.7\columnwidth]{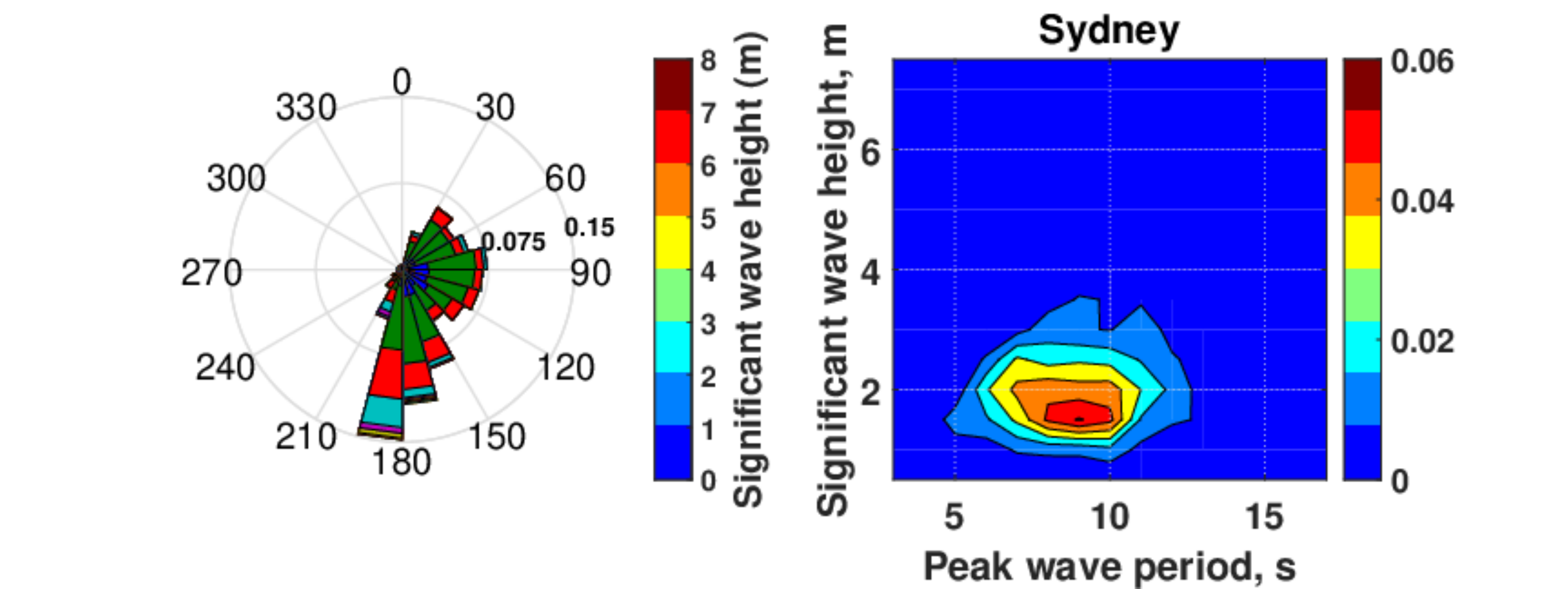}}\\
\subfloat[]{
\includegraphics[clip,width=0.7\columnwidth]{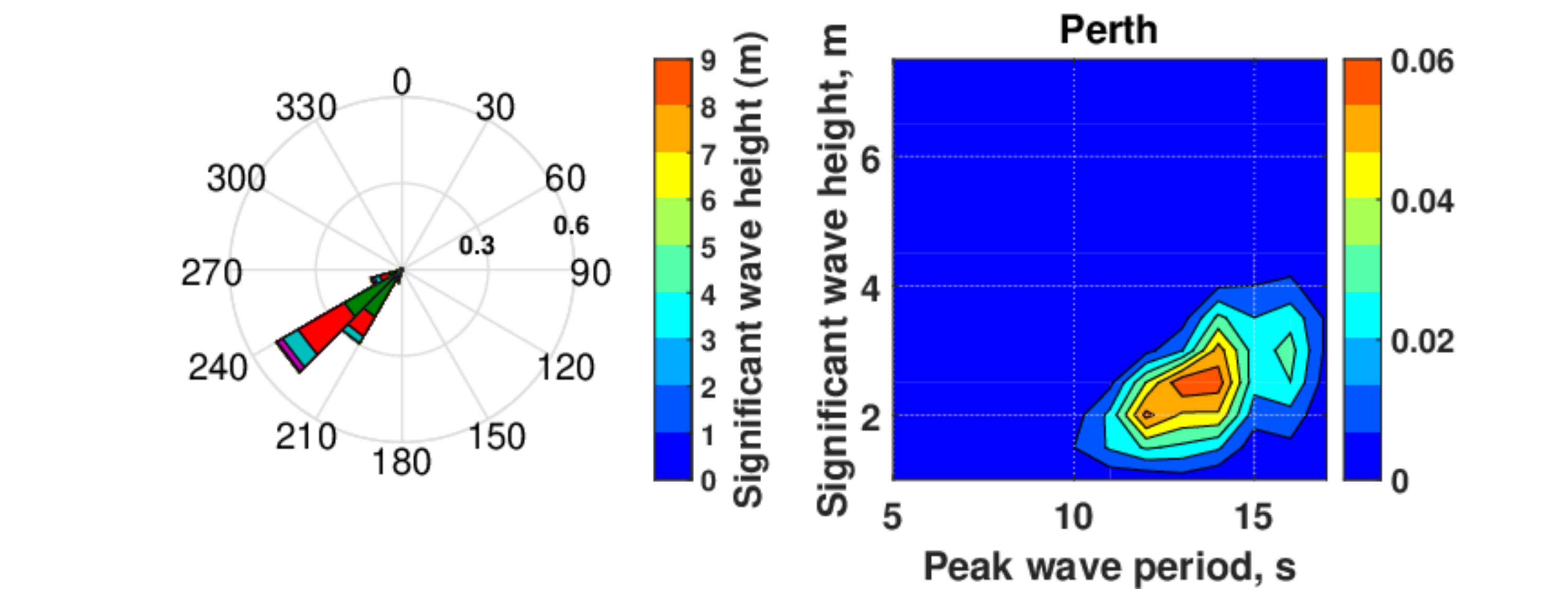}}
\caption{Wave data for two test sites in Australia: (a)~Sydney and (b)~Perth. These are: the directional wave rose (left) and wave scatter diagram (right).}%
\label{fig:wave_direct}%
\end{figure}

The total energy produced by each buoy in an array is modelled as the sum of three forces \cite{sergiienko2018three}:
\begin{enumerate}
\item The power of wave excitation ($F_{exc,p}(t)$) includes the forces of the diffracted and incident ocean waves when all generators locations are fixed. 
\item The force of radiation($F_{rad,p}(t)$) is the derived power of an oscillating buoy independent of incident waves. 
\item Power take-off force($F_{pto,p}(t)$) is the force exerted on the generators by their tethers. 
\end{enumerate}
Interactions between buoys are captured by the $F_{exc,p}(t)$ term. These interactions can be destructive or constructive, depending on buoys' relative angles, distances and surrounding sea conditions.  Equation~\ref{eqn-power} shows the power accumulating to a buoy number $p$ In a buoy array.  
\begin{equation}\label{eqn-power}
M_p\ddot{X}_p(t)=F_{exc,p}(t)+F_{rad,p}(t)+F_{pto,p}(t)
\end{equation}
Where $M_p$ is the displacement of the $p_{th}$ buoy, $\ddot{X}_p(t)$ is a vector of body acceleration in the surge, heave and sway. The last term, denoting the power take-off system, that can be simulated as a linear spring and damper. Two control factors are involved for each mooring line: the damping $B_{pto}$ and stiffness $K_{pto}$ coefficients. Therefore the Equation (\ref{eqn-power}) can be elaborated as:
\begin{equation}\label{Ex-eqn-power}
((M_{\Sigma}+A_{\sigma}(\omega))j\omega+B_{\sigma}(\omega)-\frac{K_{pto,\Sigma}}{\omega}j+B_{pto,\Sigma})\ddot{X}_{\Sigma}= \hat{F}_{exc,\Sigma}
\end{equation}
where $A_{\Sigma}(\omega))$ and $B_{\Sigma}(\omega)$ are hydrodynamic parameters which are derived from the semi-analytical model based on \cite{wu1995radiation}. Hence, the total power output of a buoy array is:
{\small
\begin{equation}\label{total-power}
P_{\Sigma}= \frac{1}{4}(\hat{F^*}_{exc,\Sigma}\ddot{X}_{\Sigma}+\ddot{X^*}_{\Sigma}\hat{F}_{exc,\Sigma})-\frac{1}{2}\ddot{X^*}_{\Sigma}B\ddot{X^*}_{\Sigma}
\end{equation}}
While we can compute the total power in Equation \ref{total-power}, it is very computationally demanding and increases exponentially with the number of buoys. With constructive interference the total power output can scale super--linearly with the number of buoys. The detailed wave characteristics including the number, direction and the probability of wave frequencies can be seen in figure~\ref{fig:wave_direct}.

\section{Optimisation Setup}\label{sec:opt}
The optimisation problem studied in this work can be expressed as:\\
\[
  P_{\Sigma}^* = \mbox{\em argmax}_{\mathbf{x,y}} P_{\Sigma}(\mathbf{x,y})
\]
\noindent ,where $P_{\Sigma}(\mathbf{x,y})$ is the average whole-farm power given by the buoys placements in a field at $x$-positions: $\mathbf{x}=[x_1,\ldots,x_N]$ and corresponding $y$ positions: $\mathbf{y}=[y_1,\ldots,y_N]$. The buoy number is here $N=16$. 
\paragraph{Constraints}
There is a square-shaped boundary constraint  for placing all buoys positions  $(x_i,y_i)$: $l\times w$ where $l=w=\sqrt{N * 20000}\,m$.
This gives \SI{20000}{\metre\squared} of the farm-area per-buoy. To maintain a safety distance, buoys must also be at least 50 metres distant from each other. For any layout $\mathbf{x,y}$ the sum-total of the inter-buoy distance violations, measured in metres, is:

\vspace{5mm}\hspace{40mm}$\mbox{\em{Sum}}_{\mbox{\em dist}}= \sum_{i=1}^{N-1}\sum_{j=i+1}^{N} 
(\mbox{\em{dist}}((x_i,y_i),(x_j,y_j))-50), $

\hspace{70mm}$\mbox{if } \mbox{\em{dist}}((x_i,y_i),(x_j,y_j))<50$ \mbox{else 0}

\noindent where $\mbox{\em dist}((x_i,y_i),(x_j,y_j))$ is the L2 (Euclidean) distance between buoys $i$ and $j$. 
The penalty applied to the farm power output (in Watts) is $(\mbox{\em{ Sum}}_{\mbox{\em{dist}}}+1)^{20}$. This steep penalty allows better handling of constraint violations during the search. Buoy placements which are outside of the farm area are handled by reiterating the positioning process.

\subsection{Computational Resources}
In this work, depending on the optimisation method,
the average evaluation time for a candidate layout
can vary greatly. To ensure a fair
comparison of methods the maximum budget for all optimisation methods 
is three days of elapsed time on a dedicated high-performance shared-memory parallel platform. The compute nodes have 2.4GHz Intel 6148 processors and 128GB of RAM. The meta-heuristic frameworks as well as the hydrodynamic simulator for $P_{\Sigma}(\mathbf{x,y})$ are run in MATLAB R2018. This MATLAB license enables us to run 12 worker threads in parallel and the methods are optimised to use as many of these threads as the methodology allows. 

\section{Methods}\label{sec:method}

In this study, the optimisation approaches employ two strategies. First, optimising all decision variables (buoy placements) at the same time. We compare five population-based  EAs that use this strategy. Second, based on \cite{neshat2018detailed}, we place one buoy at a time sequentially, comparing two hybrid techniques. 
 
\subsection{Evolutionary Algorithms (EAs)}\label{sec:allatonce}
\label{subsec:EAs}
Five popular off-the-shelf EAs are compared in the first strategy to optimise all problem dimensions. These EAs include:
 (1) Differential Evolution (DE)~\cite{storn1997differential},  with a configuration of $\lambda=30$ (population size), $F=0.5$ and $P_{cr}=0.5$; 
 (2) covariance matrix adaptation evolutionary-strategy (CMA-ES) \cite{hansen2006cma} 
with the default settings and  $\lambda=$DE configurations; 
(3) a ($\mu+\lambda$)EA  that mutates buoys' position with a
\cite{eiben2007parameter}
probability of $1/N$ using a normal distribution ($\sigma=0.1\times(U_b-L_b)$) when $\mu=50$ and $\lambda=25$; and (4) Particle Swarm optimisation (PSO) \cite{eberhart1995new}, with $\lambda$= DE configurations ,~$c_1=1.5, c_2=2,\omega=1$ (linearly decreased).

\subsection{Hybrid optimisation algorithms }\label{sec:hybrid}
Relevant researches ~\cite{neshat2018detailed,neshat2019new,Neshat:2019:HEA:3321707.3321806} noticed that employing a neighborhood search around the previously placed-buoys could be beneficial for exploiting constructive interactions between buoys. The two following methods utilise this observation by placing and optimising the position of one buoy at a time.

\subsubsection{Local Search + Nelder-Mead(LS-NM)}
LS-NM ~\cite{neshat2018detailed} is one of the most effective and fast WEC placement methods. LS-NM positions generators sequentially by sampling at a normally-distributed random deviation ($\sigma=$\SI{70}{\metre}) from the previous buoy location.  
The best-sampled location is optimised using $N_s$ iterations of the Nelder-Mead search method. This process is repeated until all buoys are placed.


\begin{figure}[t]%
\centering
\includegraphics[width=0.9\textwidth]{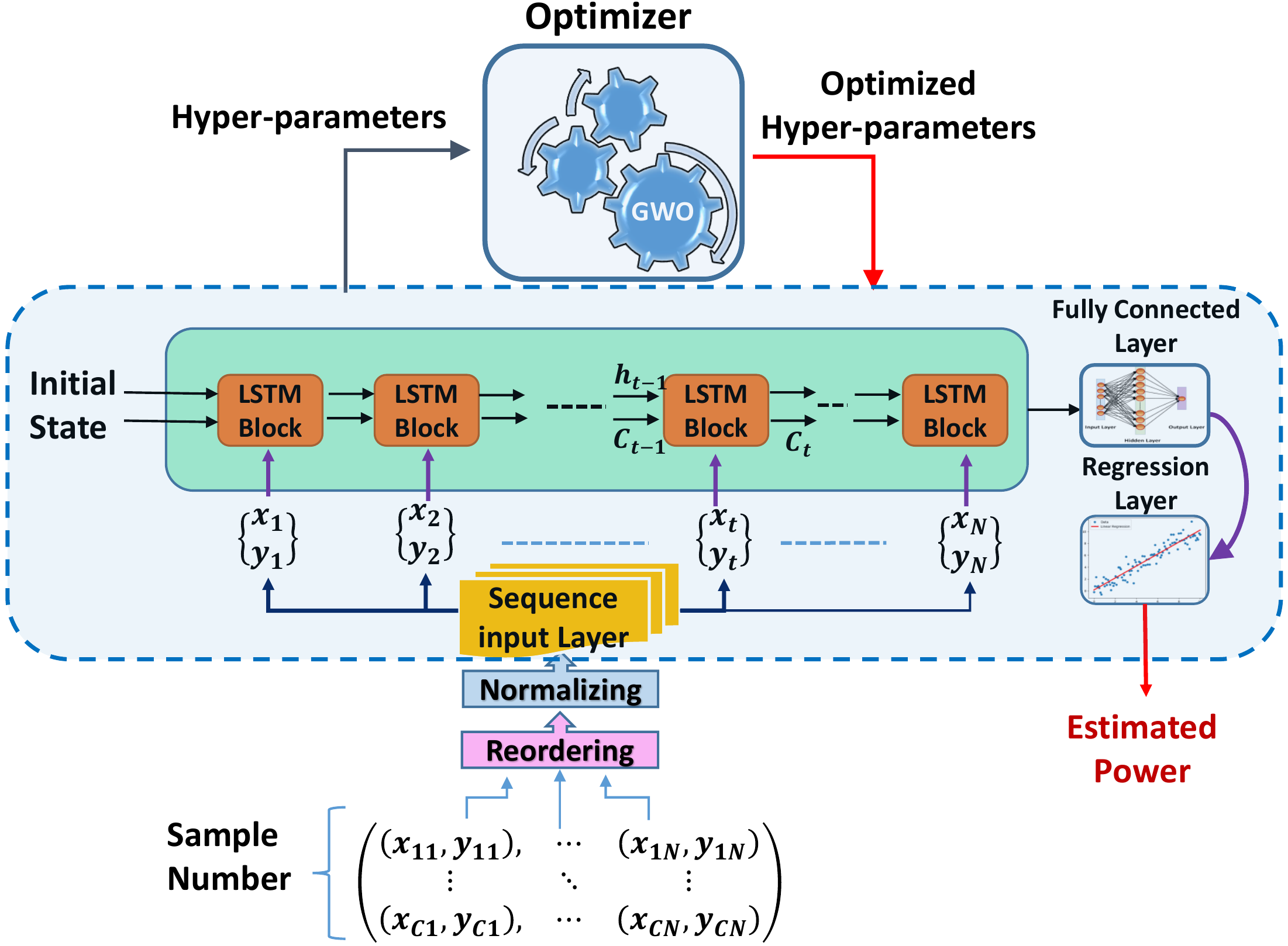}
\caption{The Neuro-Surrogate model architecture}
\label{fig:LSTM}
\end{figure}
 \subsubsection{Adaptive Neuro-Surrogate Optimisation method (ANSO)}
Given the complexity of the optimisation problem we devise a novel approach with the intuition that (a) sequential placement of the converters provide a simple, yet effective baseline and (b) we can \emph{learn} a surrogate to mimic the potential power output for an array of buoys.
Hence, we provide a three step solution (as detailed in Algorithm~\ref{alg:ANSO}).

 
\begin{algorithm}[t]
\caption{$\mathit{ANSO}$}\label{alg:ANSO}
\label{alg:anso}
\begin{algorithmic}[1]
\Procedure{Adaptive Neuro-Surrogate Optimisation (ANSO)}{}\\
 \textbf{Initialization}
 \\$\mathit{size}=\sqrt{N*20000}$  \Comment{Farm size}
 \\$\mathit{Res}=3$  \Comment{angle resolution}
  \\$\mathit{angle=\{0,Res,2\times Res,\ldots,360-Res\}}$ \Comment{symmetric samples angle}
 \\ $\mathit{iters}=Size([angle])$ \Comment{Number of symmetric samples}
 \\ $\mathit{EvalSet}=\{2^{nd},3^{rd},5^{th},...,15^{th} \}$ \Comment{Set of evaluated buoys}
 \\ $\mathit{EstimSet}=\{4^{th},6^{th},8^{th},...,16^{th} \}$ \Comment{Set of estimated buoys}
\\$\mathit{S}=\{\langle x_1,y_1 \rangle,\ldots,\langle x_N,y_N \rangle\}=\bot$ \Comment{Positions}
\\ $\mathit{S}_{(1)}=\{\langle size,0\rangle\}$  \Comment{first buoy position}
\\ $\mathit{energy}=\mathit{Eval}([S_{(1)}])$
\\$\mathit{bestPosition}=S_{(1)}$; 
 \For{ $i$ in $[2,..,N]$ }
\State$\mathit{bestEnergy}=0;$

\If{$\mathit{i \in Evalset}$} \Comment{ layouts should be evaluated by Simulator}

 \For{$j$ in $[1,..,\mathit{iters}]$}
\State $(Sample_{j},\mathit{energy_{j}})$={\em{SymmetricSampleEval}}$(\mathit{angle_j},S_{(i-1)})$ 
    \If{ {\em{$Sample_j$ is feasible $\&$ $energy_j$ $>$ bestEnergy}} }
       \State  $\mathit{tPos}=\mathit{Sample_j}$ \Comment{ Temporary buoy position}
      \State $\mathit{bestEnergy}=\mathit{energy_j}$
      \State $\mathit{bestAngle}=\mathit{j}$
  \EndIf
  
  \EndFor
   
  \If{ {\em{No feasible solution is found}} }
      \State $(Sample_{1},\mathit{energy_{1}})$={\em{rand($\mathit{S_{(i-1)}}$)}} 
 \EndIf
   \State $(Es_1,Es_2)$={\em{SymmetricSampleEval}}$(\mathit{bestAngle\pm Res/2},S_{(i-1)})$ 
   \State $(S_{(i)},\mathit{energy})$={\em{FindbestS}}$(\mathit{tPos},Es_1,Es_2)$ 
   \State $\mathit{DataSet_i}=\mathit{UpdateData(Sample,energy)}$
   
      \Else \Comment{ layouts should be estimated by the LSTM}
       \State $(\mathit{HyperParameters_i})$={\em{Optimise-Hyper}}$(\mathit{DataSet_i})$ \Comment{Optimising by GWO}
   \State $(\mathit{Deep_i})$={\em{reTrain}}$(\mathit{Deep_i,DataSet_i,HyperParameters_i})$
       \For{$j$ in $[1,..,\mathit{iters}]$}
\State $(Sample_{j},\mathit{energy_{j}})$={\em{SymmetricSampleEstim}}$(\mathit{angle_j},S_{(i-1)},\mathit{Deep_i})$ 
    \If{ {\em{$Sample_j$ is feasible $\&$ $energy_j$ $>$ bestEnergy}} }
       \State  $\mathit{tPos}=\mathit{Sample_j}$ \Comment{ Temporary buoy position}
      \State $\mathit{bestEnergy}=\mathit{energy_j}$
      \State $\mathit{bestAngle}=\mathit{j}$
  \EndIf
  
  \EndFor
  
    \EndIf
  \EndFor 
  \State $(\mathit{bestPosition'},\mathit{bestEnergy'})=\mathit{BackTrackingOp}(\mathit{bestPosition})$  
\State \textbf{return} $\mathit{bestPosition'},\mathit{bestEnergy'}$  \Comment{Final Layout}
\EndProcedure
\end{algorithmic}
\end{algorithm}


\paragraph{Symmetric Local Search (SLS):}

Inspired by LS-NM ~\cite{neshat2018detailed,Neshat:2019:HEA:3321707.3321806}, in the first step we sequentially place buoys by conducting a local search for each placement. SLS 
starts by placing the first buoy in the recommended position (bottom corner of the field)
and then for each subsequent buoy position, uniformly performs $N_{iters}$ of feasible local samples are made in different sectors commencing at angles: $\{angles=[\ang{0}, Res^\circ, 2\times Res^\circ,...,360-Res^\circ]\}$ 
and bounded by a radial distance of between $50+R_{1}m$ (safe distance+first search radius ) and $50+R_2m$. 
The best sample is chosen among the $N_{iters}$ local samples. Next, two extra neighbourhood samples near the best sample ($\pm Res/2$) are made for increasing the exploitation ability of the method. The best of these three samples, based on total absorbed power, is then selected.

\paragraph{Learning the neuro-surrogate model:}
The hydrodynamic simulator is computationally expensive to run. 
A fast and accurate neuro-surrogate is used here
to estimate the power of layout based on the position of the next buoy: ($x_i,y_i$).
Our motivation is that a fast surrogate function  can quickly estimate what the simulator takes a long time to compute.
The key challenges to overcome in designing a neuro-surrogate are:  
1) {\em{function complexity}}: a highly nonlinear and complex relationship between buoys position and absorbed farm power, 
2) {\em{changing dataset}}: as more evaluations of the placements are performed, new data for training is collected that has to be incorporated, and, 3){\em{efficiency}}: training time plus the hyper-parameter tuning has to be included in our computational budget. 

For handling these challenges, we use a combination of recurrent networks with LSTM cells~\cite{hochreiter1997long} (sequential learning strategy), and, an optimiser (GWO)~\cite{mirjalili2014grey} for tuning the network hyper-parameters for estimating the power of the layouts. 
The overall framework is shown in Figure~\ref{fig:LSTM}. The proposed LSTM network is designed for sequence-to-one regression in which the input layer is from 2D buoy positions ($x_i,y_i$) and the output of the regression layer is the estimated layout power.  
The LSTM training process is done using the back-propagation algorithm, in which the gradient of the cost function (in this case the mean squared error between the true ground-truth and the estimated output of the LSTM) at different layers are computed to update the weights. 


\begin{table*}[t]

\scalebox{0.6}{
\begin{tabular}{l|l|l|l|l|l|l|l|l|l|l|l|l|l}
\hlineB{4}
\multicolumn{13}{ c }{\textbf{\begin{large}Perth wave scenario (16-buoy)\end{large}}} \\
\hlineB{4}
                    & \textbf{DE  }        & \textbf{CMA-ES}      & \textbf{PSO}         & $(\mu+\lambda)$ \textbf{EA}   & \textbf{LS-NM}       & \textbf{ANSO-$S_1$}     & \textbf{ANSO-$S_2$}     & \textbf{ANSO-$S_3$}     & \textbf{ANSO-$S_4$}     & \textbf{ANSO-$S_1$-B}  & \textbf{ANSO-$S_2$-B}  & \textbf{ANSO-$S_3$-B}  & \textbf{ANSO-$S_4$-B } \\
                    \hlineB{4}
\texttt{\textbf{Max}} & 1474381 & 1490404 & 1463608 & 1506746 & 1501145 & 1544546 & 1533055 & 1554926 & 1555446 & 1552108 & 1549299 & 1554833 & \textbf{1559535}  \\
\hlineB{2}
\texttt{\textbf{Min}} & 1455256 & 1412438 & 1433776 & 1486397 & 1435714 & 1513894 & 1489365 & 1531290  & 1543637 & 1535508 & 1502373 & 1543384 & 1549517  \\
\hlineB{2}
\texttt{\textbf{Mean}}  & 1462331 & 1476503 & 1450589 & 1494311 & 1479345 & 1534032 & 1514147 & 1543361 & 1550171 & 1544832 & 1525112 & 1549276  & \textbf{1556073} \\
\hlineB{2}
\texttt{\textbf{Median}}& 1462697  & 1482974  & 1448835 & 1493109 & 1490195  & 1535162 & 1516162 & 1544076 & 1551105 & 1544733 & 1523082 & 1549701 & \textbf{1556091} \\
\hlineB{2}
\texttt{\textbf{STD}} & 4742.1 & 23004.6 & 8897.7 & 6227.9 & 23196.4  & 7991.1 & 12092.2 & 7441.2 & 3333.5 & 5531.3  & 12663.7 & 4006.2 & 2783.2 \\

\hlineB{4}
\multicolumn{13}{ c }{\textbf{\begin{large}Adelaide wave scenario (16-buoy)\end{large}}} \\
\hlineB{4}
\texttt{\textbf{Max}} & 1494124 & 1501992 & 1475991 & 1517424 & 1523773  & 1563935 & 1563249 & 1583623 & 1585626 & 1576713 & 1571181 & \textbf{1589830} & 1588297 \\
\hlineB{2}
\texttt{\textbf{Min}}  & 1468335 & 1478052 & 1452804 & 1488276 & 1496878 & 1558613  & 1520681 & 1565725 & 1571131 & 1566240 & 1527665 & 1567491 & 1576009  \\
\hlineB{2}
\texttt{\textbf{Mean}} & 1479247 & 1488783  & 1461579 & 1502708 & 1513070 & 1561624 & 1541404 & 1573125 & 1575439 & 1572454 & 1552201 & \textbf{1581643} & 1578365 \\
\hlineB{2}
\texttt{\textbf{Median}}& 1479707 & 1487430 & 1460687 & 1501805 & 1515266 & 1562548 & 1541101 & 1576658 & 1575092 & 1573763 & 1552663 & \textbf{1582515} & 1577353 \\
\hlineB{2}
\texttt{\textbf{STD}} & 7704.9 & 8167.9 & 6670.9  & 8443.2  & 7434.7 & 2154.3 & 12366.9 & 7572.5 & 3676.1 & 3639.8 & 12373.1 & 6481.1 & 3428.9\\

\hlineB{4}
\multicolumn{13}{ c }{\textbf{\begin{large}Sydney wave scenario (16-buoy)\end{large}}} \\
\hlineB{4}
\texttt{\textbf{Max}}  & 1520654 & 1529809 & 1525938  & 1528934 & 1524164 & 1523552 & 1523353 & 1523549 & 1524974 & 1531566 & \textbf{1532200} & 1528619 & 1531155 \\
\hlineB{2}
\texttt{\textbf{Min}}  & 1515231 & 1520031 & 1508729 & 1516014 & 1487836 & 1509677 & 1493596 & 1500115 & 1514248 & 1517559 & 1506128 & 1513182 & 1520086 \\
\hlineB{2}
\texttt{\textbf{Mean}} & 1518047 & 1524054 & 1519251 & 1522625 & 1507594 & 1517627 & 1514384  & 1514300 & 1520597  & 1524357  & 1523382 & 1521277 & \textbf{1526443} \\
\hlineB{2}
\texttt{\textbf{Median}} & 1518014 & 1523440  & 1520319 & 1522234 & 1507898 & 1518667 & 1516523  & 1518055 & 1521351  & 1524767 & 1524356 & 1522289 & \textbf{1527839} \\
\hlineB{2}
\texttt{\textbf{STD}} & 1880.1 & 2767.8 & 5818.1 & 3887.1 & 10929.2 & 4871.7 & 8811.3 & 8642.02 & 4021.8 & 4161.7 & 6710.5 & 6393.02 & 3481.1\\

\hlineB{4}
\multicolumn{13}{ c }{\textbf{\begin{large}Tasmania wave scenario (16-buoy)\end{large}}} \\
\hlineB{4}
\texttt{\textbf{Max}}  & 3985498 & 4063049 & 3933518  & 4047620 & 4082043 & 4144344  & 4085915 & 4121312 & 4135256  & \textbf{4162505} & 4104237 & 4143536 & 4160738  \\
\hlineB{2}
\texttt{\textbf{Min}} & 3935990 & 3935833 & 3893456 & 3992362 & 3904892 & 4025709 & 4021772 & 4071497 & 4113146 & 4053715 & 4043849  & 4103441 & 4128702 \\
\hlineB{2}
\texttt{\textbf{Mean}}  & 3956691 & 4000087 & 3914316 & 4019472 & 4008228 & 4072874  & 4042537 & 4093453 & 4122447 & 4095608 & 4071852  & 4123334  & \textbf{4145569} \\
\hlineB{2}
\texttt{\textbf{Median}}& 3951489 & 3994739 & 3914764 & 4019623 & 4020515 & 4066904 & 4033063 & 4091620 & 4121959 & 4079286 & 4074154 & 4124520 & \textbf{4144359} \\
\hlineB{2}
\texttt{\textbf{STD}}  & 17243.1 & 37701.2 & 13758.4 & 18377.5 & 54771.9  & 33897.8 & 19819.9 & 17367.4 & 6422.9 & 34789.9 & 16516.9 & 12411.4 & 10085.3\\
\hlineB{4}
\end{tabular}
}
\caption{Performance comparison of various heuristics for the 16-buoy case, based on maximum, median and mean power output layout of the best solution per experiment.}
\label{table:allresults16}
\end{table*}
For tuning the hyper-parameters of the LSTM we use the ranges: MiniBatch size ($5-100$), learning rate ($10^{-4}-10^{-1}$), the maximum number of training epochs ($50-600$), LSTM layer number (one or two) and hidden node number ($10-150$). At each step of the position optimisation, a fast and effective meta-heuristic algorithm (GWO) ~\cite{mirjalili2020grey} is used. This is because the collected data-set is dynamic in terms of input length (increases over time) and the arrangement of buoys. This hybrid idea depicts an adaptive learning process that is fast (is converged by a few evaluations (Figure~\ref{fig:hyper_optimization})), accurate and easily scalable to larger sizes.

\paragraph{Backtracking Optimisation:}
The third component of ANSO is applying a backtracking optimisation strategy (BO). This is because the initial placements described above are based on greedy selection, the previous buoys' positions are revisited during this phase. Consequently, introducing backtracking can help maximise the power of the layouts. For this part, a 1+1EA\cite{droste2002analysis} is employed. In each iteration, the buoys position $(x_i,y_i)$ is mutated based on a Gaussian normally distributed  random  variable  with a dynamic mutation step size ($\sigma$) that is decreased linearly and an adaptive probability rate ($P_m$). The mutated position is evaluated by the simulator. Both Equations~\ref{eq:sigma} and~\ref{eq:ProMu} represent the details of these control parameters of the BO method.  

\begin{equation}
\label{eq:sigma}
 \sigma_{iter}=\sigma_{Max} \times 0.08 \times iter/ iter_{Max} ~\forall ~ iter \in \{1,...,iter_{Max}\}
\end{equation}
\begin{equation}
\label{eq:ProMu}
 P_{m_i}=(1/N)\times (1 / (Power_{Buoy_i}/Max_{Power})) + \omega_i ~\forall~ i \in \{1,...,N\}
\end{equation}
Where $\sigma_{Max}$ is the initial mutation step size at \SI{10}{\metre} and $P_{m_i}$ shows the mutation probability of each buoy in the layout. We assume that the buoys with lower absorbed power need more chance of modification, so the highest mutation probability rate should be allocated to that buoy with the lowest power and vice-versa. In addition, $\omega_i$ is a weighted linear coefficient  from $0.1$ (for the lowest power of the buoys) to $0$ (highest buoy power). The reserved runtime for the BO method is one hour. Algorithm \ref{alg:ANSO} describes this method in detail. 
\begin{figure}[tb]
\centering
\includegraphics[width=\textwidth]
{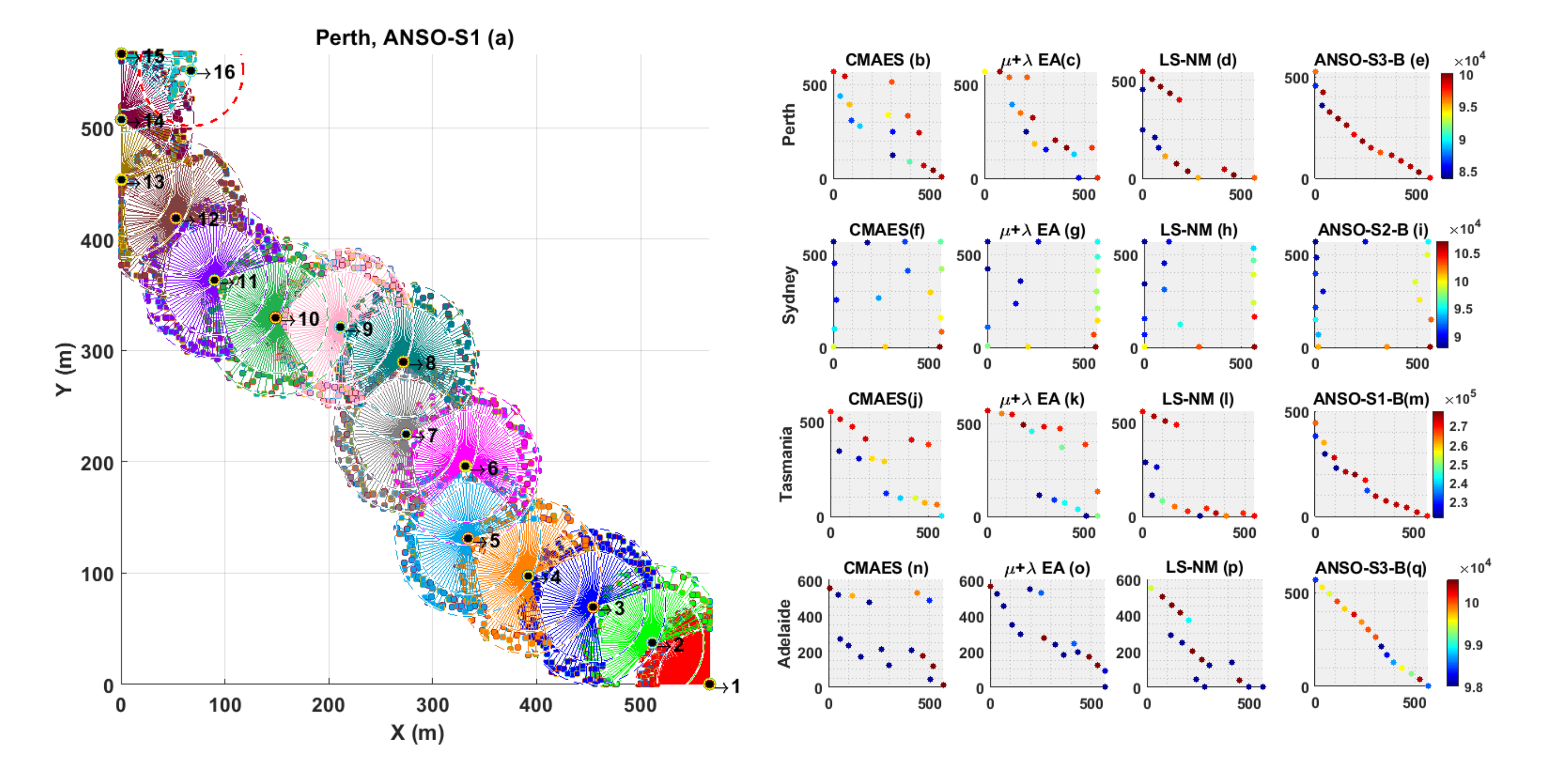}

\caption{The best-obtained 16-buoy layouts: figure(a) presents how the proposed hybrid method can optimise buoys position and estimate the power of some buoys ($4^{th},6^{th},...,16^{th}$) sequentially. Two rings around each buoy show the exploration space($Res=3^\circ$,$\Delta R=$\SI{20}{\metre}). Other figures show the best layouts arrangement of the four real wave scenarios based on Table \ref{table:allresults16}.}\label{fig:best_solutions}
\end{figure}

\section{Experiments}\label{sec:experiments}
The adaptive tuning of hyper-parameters in ANSO makes it
compatible with each layout problem. 
Moreover, no pre-processing time is required for collecting the relevant training data-set, ANSO is able to collect the required training data in real-time during the sampling and optimisation of previous buoy positions. 

In this work, we test four strategies for buoy placement under
ANSO these vary in the membership of the $\mathit{EvalSet}$ -- the set of buoys evaluated by the full simulator and the sample set used to train the neuro-surrogate model. The strategies tested are:
1) With $\mathit{EvalSet}=\{2^{nd},3^{rd},5^{th},...,15^{th} \}$
so the neuro-surrogate is used to place buoys: $4,6,8,\ldots,16$. The neuro-surrogate is trained prior to each placement using sampled positions used for the previous buoy placement.
2) Use $\mathit{EvalSet}=\{2^{nd},3^{rd},6^{th},9^{th},...,15^{th} \}$
so the neuro-surrogate is used to place buoys: $4,5,7,8,10,11,13,14,16$. As with strategy 1 above, the previous sampled positions evaluated by the simulator are used to train
the neuro-surrogate model.
3) The same setup as the first strategy but the LSTM is trained by {\em{all}} previous simulator samples.  
4) All evaluations are done by the simulator. 


\begin{table}
\caption{The average ranking of the proposed methods for $\rho$ by Friedman test.}
\label{table:Friedman}
\begin{tabular}{l|l|l|l|l}
\hlineB{4}
  \textbf{Rank} & \textbf{Adelaide} & \textbf{Perth}              & \textbf{Sydney}             & \textbf{Tasmania} \\ \hlineB{4}
\textbf{1 } & \textbf{ANSO-$S_3$-B~~~(1.75)}&\textbf{ANSO-$S_4$-B~~~(1.08)}&\textbf{ANSO-$S_4$-B~~~(3.00)} & \textbf{ANSO-$S_4$-B~~~(1.25) }\\ \hlineB{2}

\textbf{2 } & ANSO-$S_4$-B\quad(2.08) & ANSO-$S_4$\qquad(3.08)& ANSO-$S_1$-B\quad(4.17) & ANSO-$S_3$-B\quad(2.75) \\ \hlineB{2}

\textbf{3}& ANSO-$S_3$\qquad(3.67)& ANSO-$S_3$-B\quad(3.17) & CMA-ES~~~~\quad(4.33) & ANSO-$S_4$\qquad(3.00)       \\ \hlineB{2}

\textbf{4 } & ANSO-$S_4$\qquad(3.67) & ANSO-$S_1$-B\quad(4.00)& ANSO-$S_2$-B\quad(4.50) & ANSO-$S_1$-B\quad(4.67) \\ \hlineB{2}

\textbf{5}& ANSO-$S_1$-B\quad(4.00)& ANSO-$S_3$\qquad(4.42)& ANSO-$S_3$-B~\quad(5.08) & ANSO-$S_3$\qquad(5.00)       \\ \hlineB{2}

\textbf{6 }& ANSO-$S_1$\qquad(6.08) & ANSO-$S_1$\qquad(6.00)& $(\mu+\lambda)$EA\qquad(6.00)& ANSO-$S_2$-B\quad(6.00)    \\ \hlineB{2}

\textbf{7 }& ANSO-$S_2$-B\quad(6.75) & ANSO-$S_2$-B\quad(6.50)& ANSO-$S_4$\qquad(7.08) & ANSO-$S_1$\qquad(6.33)    \\ \hlineB{2}

\textbf{8} & ANSO-$S_2$\qquad(8.08)& ANSO-$S_2$\qquad(8.00)& PSO~\qquad\qquad(7.92) & ANSO-$S_2$\qquad(8.33)    \\ \hlineB{2}

\textbf{9} & LS-NM\qquad\quad(9.17) & $(\mu+\lambda)$EA\qquad(9.25)& ANSO-$S_1$\qquad(8.42)& LS-NM~~~\qquad(9.25)       \\ \hlineB{2}

\textbf{10}& $(\mu+\lambda)$EA~\quad(9.92)& LS-NM~~~\qquad(10.50)& DE~~\qquad\qquad(9.58) & $(\mu+\lambda)$EA\qquad(9.42)   \\ \hlineB{2}

\textbf{11}& CMAES~\qquad(11.00) & CMA-ES~\qquad(10.67)& ANSO-$S_3$\qquad(9.67)& CMA-ES\qquad(10.25)     \\ \hlineB{2}

\textbf{12}& DE~\qquad\qquad(11.83)& DE~~\qquad\qquad(11.83)& ANSO-$S_2$\qquad(10.00)& DE~~\qquad\qquad(11.75)         \\ \hlineB{2}

\textbf{13}& PSO~~\quad\qquad(13.00)& PSO~~~\quad\qquad(12.50)& LS-NM\quad\qquad(11.25)& PSO~~\qquad\quad(13.00)          
\end{tabular}
\end{table}
\begin{figure}[t]
\centering
\subfloat[]{
\includegraphics[clip,width=0.5\columnwidth]{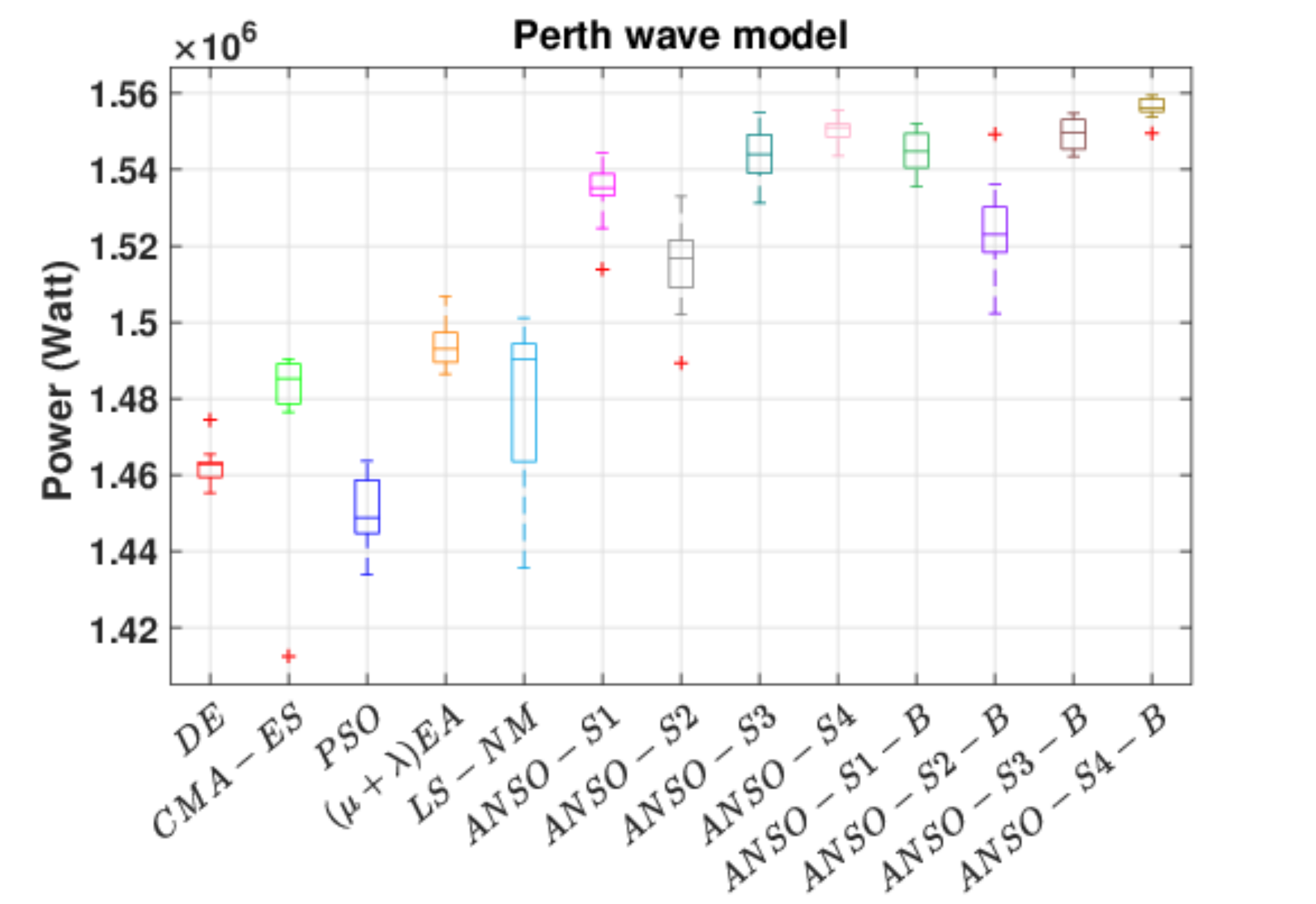}}
\subfloat[]{
\includegraphics[clip,width=0.5\columnwidth]{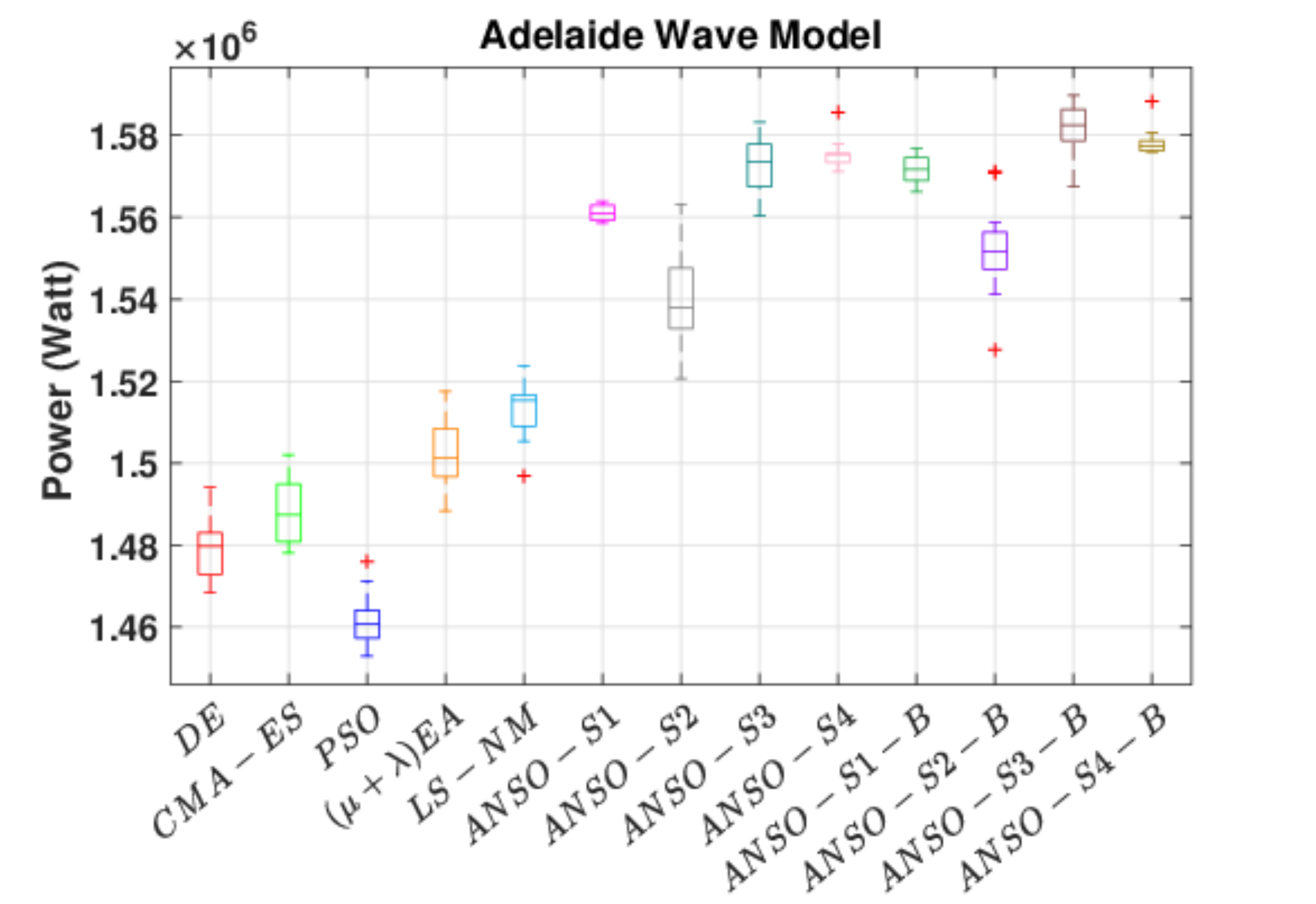}}\\
\subfloat[]{
\includegraphics[clip,width=0.5\columnwidth]{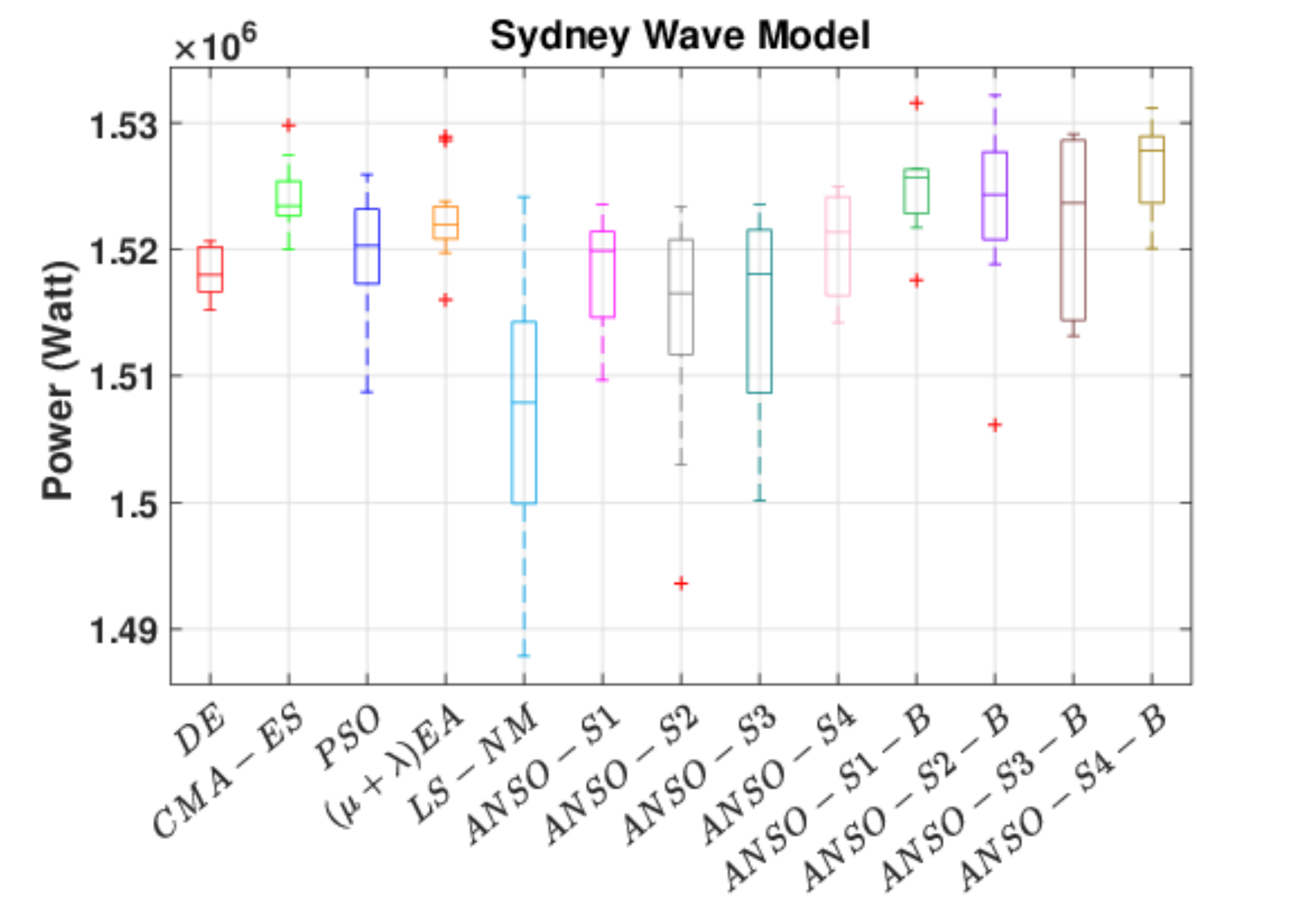}}
\subfloat[]{
\includegraphics[clip,width=0.5\columnwidth]{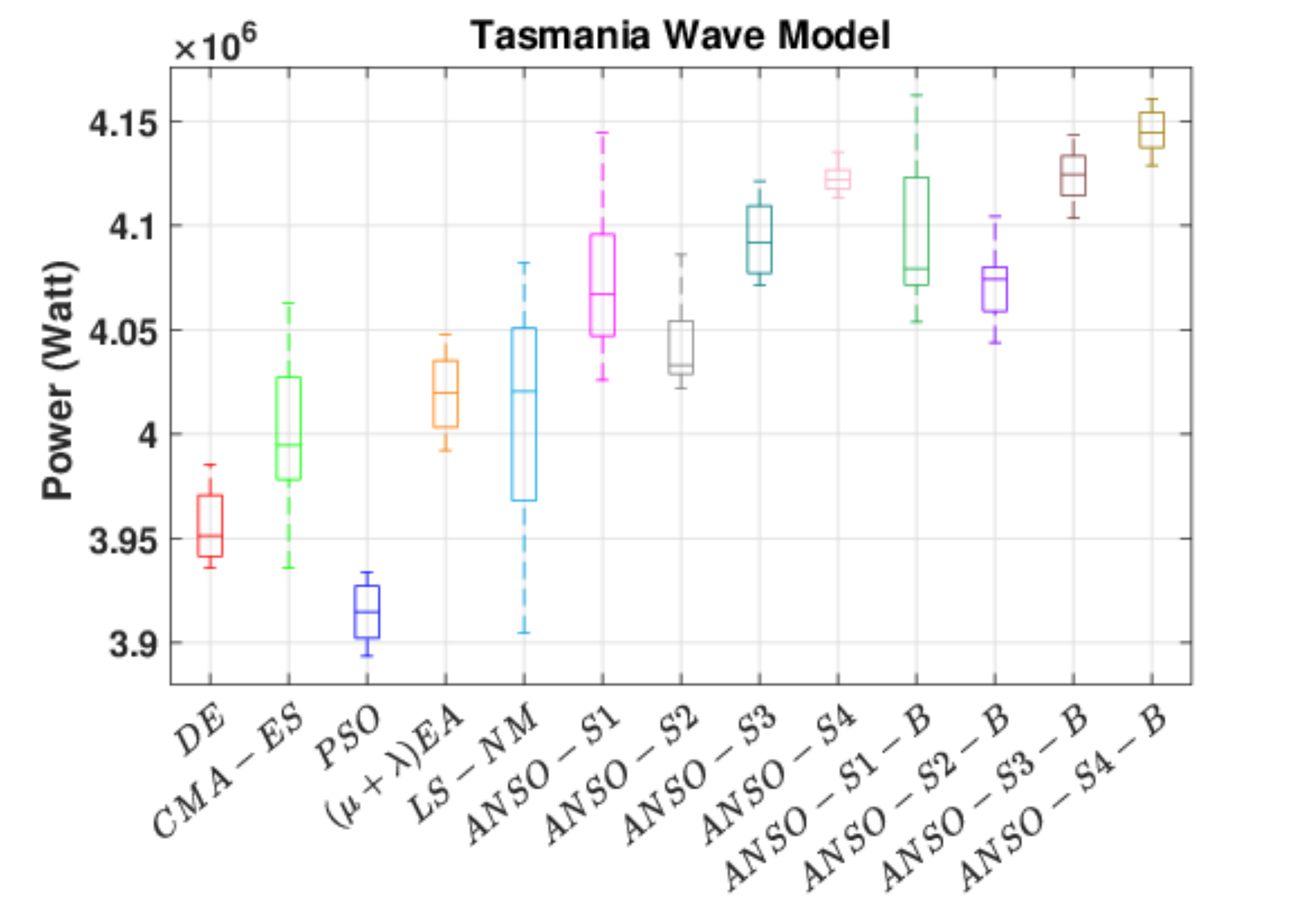}}
\caption{Comparison of the  algorithms' effectiveness for 16-buoy layouts in four real wave scenarios (Perth(a), Adelaide(b), Sydney(c), Tasmania(d)). The optimisation results show the best solutions of 10 independent runs.}%
\label{fig:box_plot}
\end{figure}

First, Table~\ref{table:allresults16}  shows the statistical results of the maximal power output of the 13 compared heuristics for 10 independent runs each for four real wave scenarios. As shown in Table~\ref{table:allresults16}, ANSO-$S_4$-B is best, on average, in Sydney, Perth and Tasmania. ANSO-$S_3$-B shows the best performance in Adelaide. However, all methodologies using the neuro-surrogate are competitive in terms of performance. 
The results of applying the Friedman test are shown in Table \ref{table:Friedman}. Algorithms are ranked according to their best configuration for each run. Again, ANSO-$S_3$-B obtained first ranking in the Adelaide wave model and non-neuro-surrogate ANSO-$S_4$-B algorithm ranks highest in other scenarios. 
The best 16-buoy layouts of the 4 compared algorithms (CMAES, ($\mu+\lambda$)EA, LS-NM and the best-performing versions of ANSO) are shown in Figure~\ref{fig:best_solutions}.  The sampling used by the optimisation process of ANSO-$S_1$ is shown in Figure~\ref{fig:best_solutions}(a). It shows how  ANSO-$S_1$  explores each buoy's neighbourhood and modifies positions during backtracking.

Figure~\ref{fig:box_plot} shows box-and-whiskers plots for the best solutions power output per run for all approaches and all wave scenarios. It can be seen that the best mean performance is given by ANSO-$S_4$-B in three of four-wave scenarios. In the Adelaide case study, ANSO-$S_3$-B performs best. Another interesting observation is that, among population-based EAs, ($\mu+\lambda$)EA excels. However, both ANSO and LS-NM outperform all population--based methods.

\begin{figure}[t]
  \centering
  \includegraphics[height=7cm]{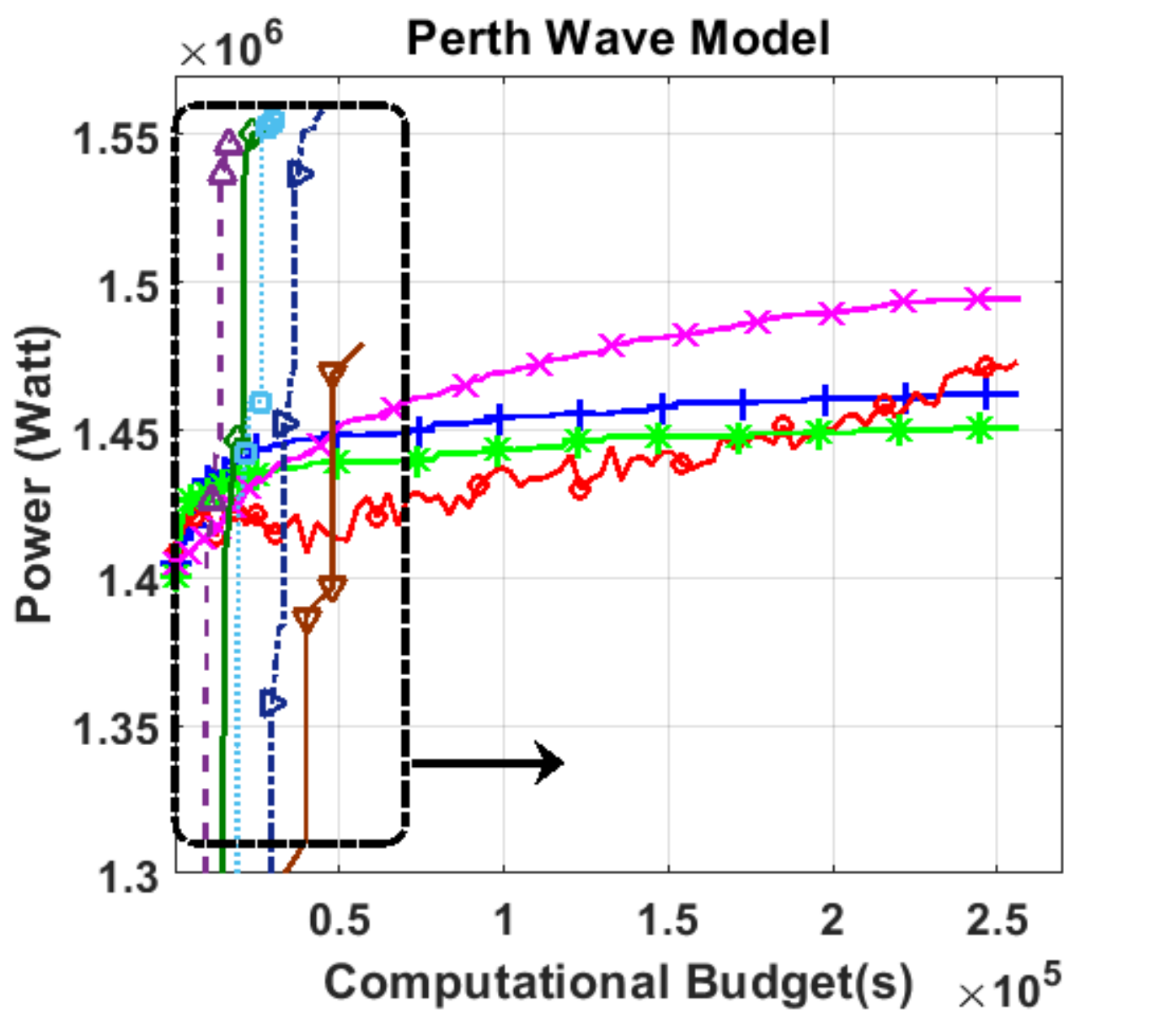}\llap{\raisebox{0.8cm}{\includegraphics[height=3.2cm]{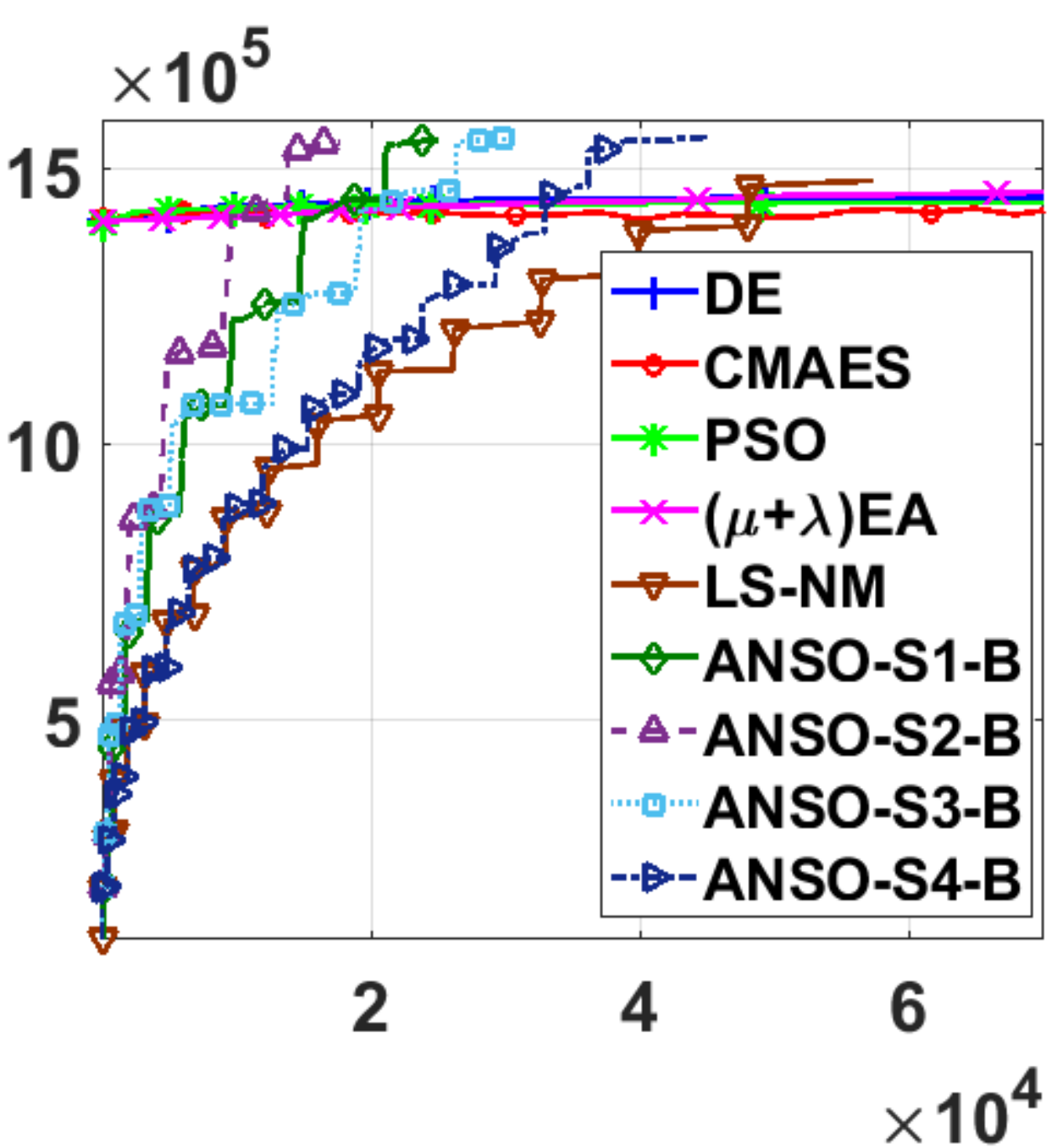}}}
  \includegraphics[height=7cm]{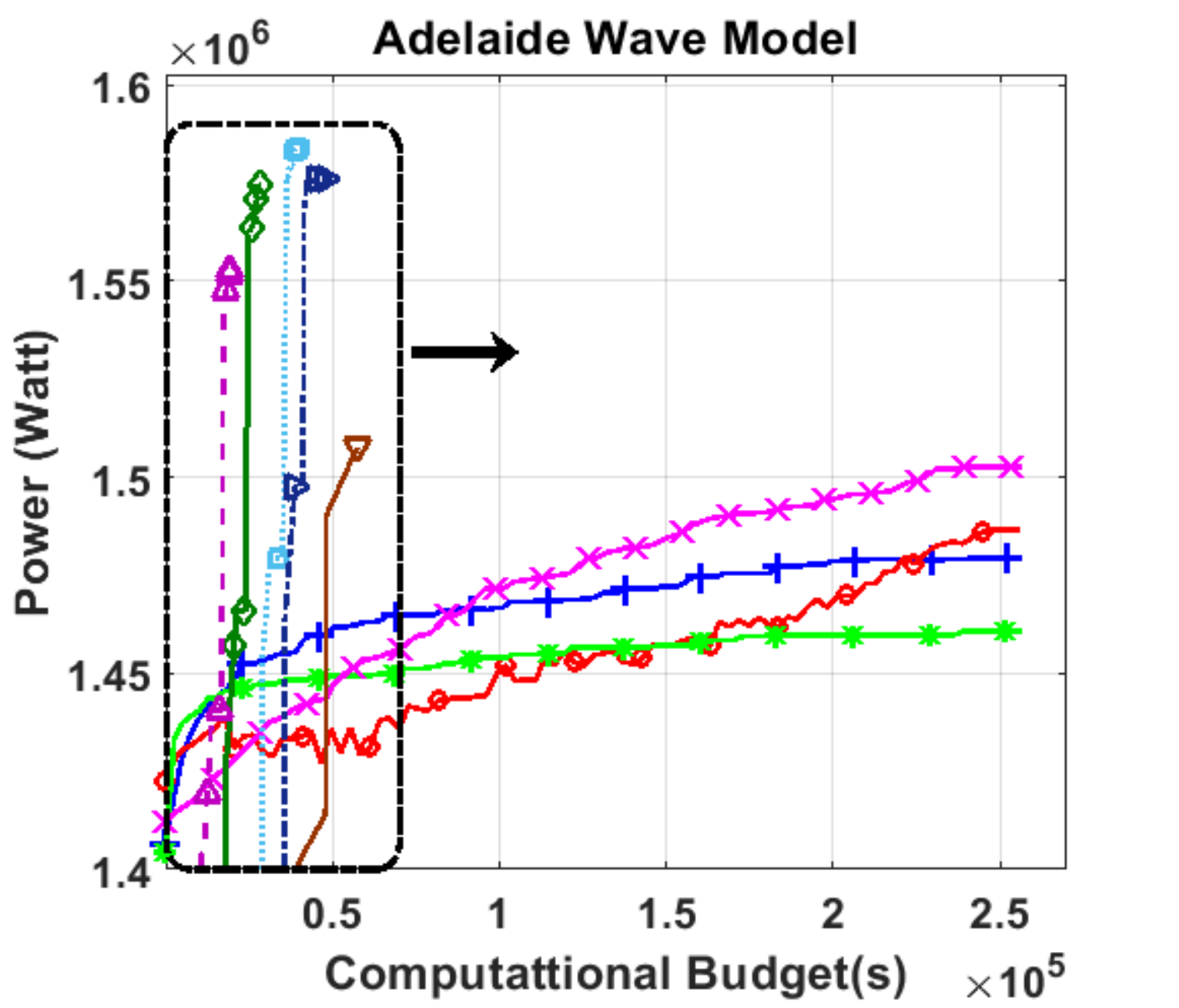}\llap{\raisebox{3.8cm}{\includegraphics[height=2.85cm]{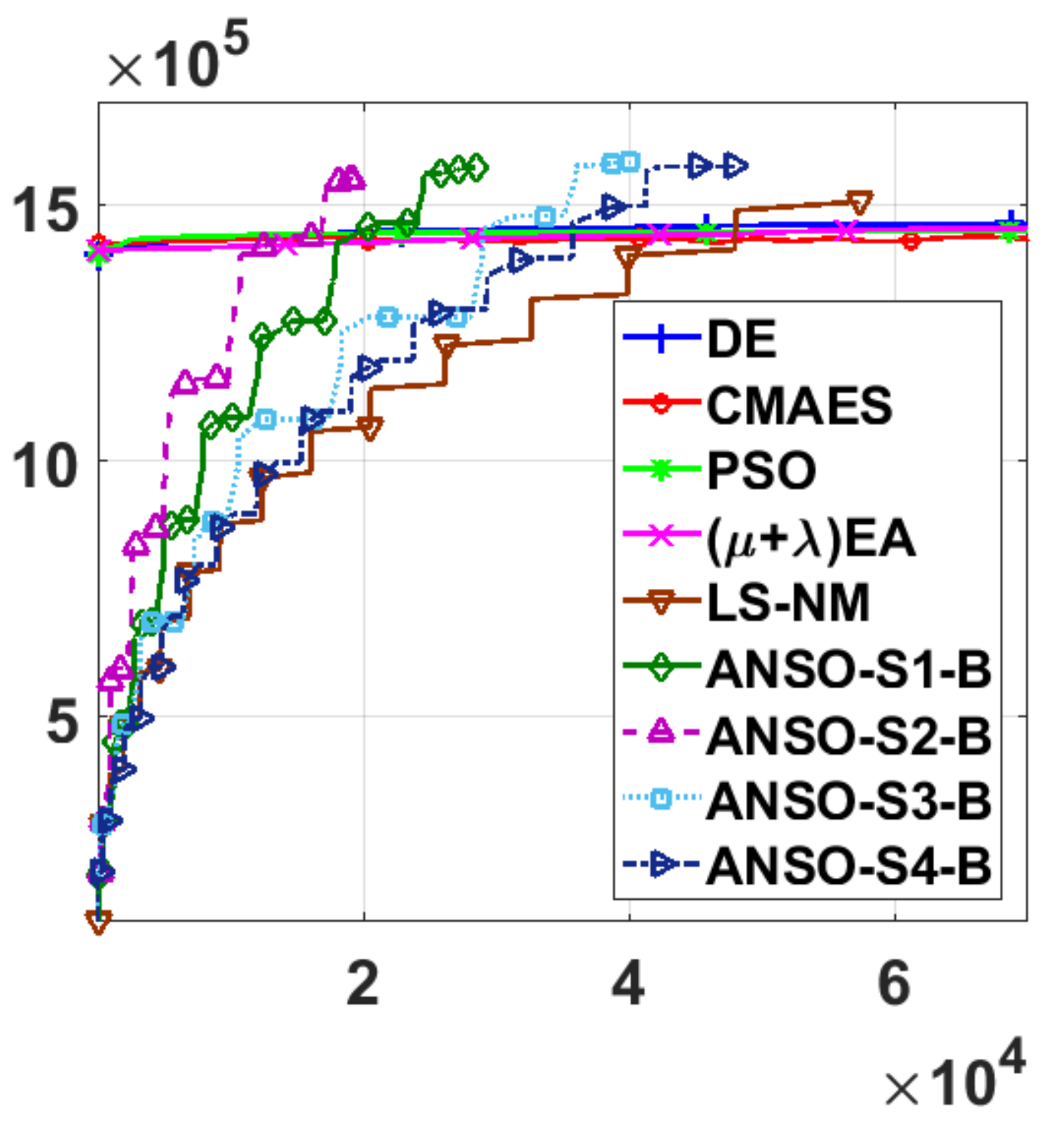}}}\\
  \includegraphics[height=7cm]{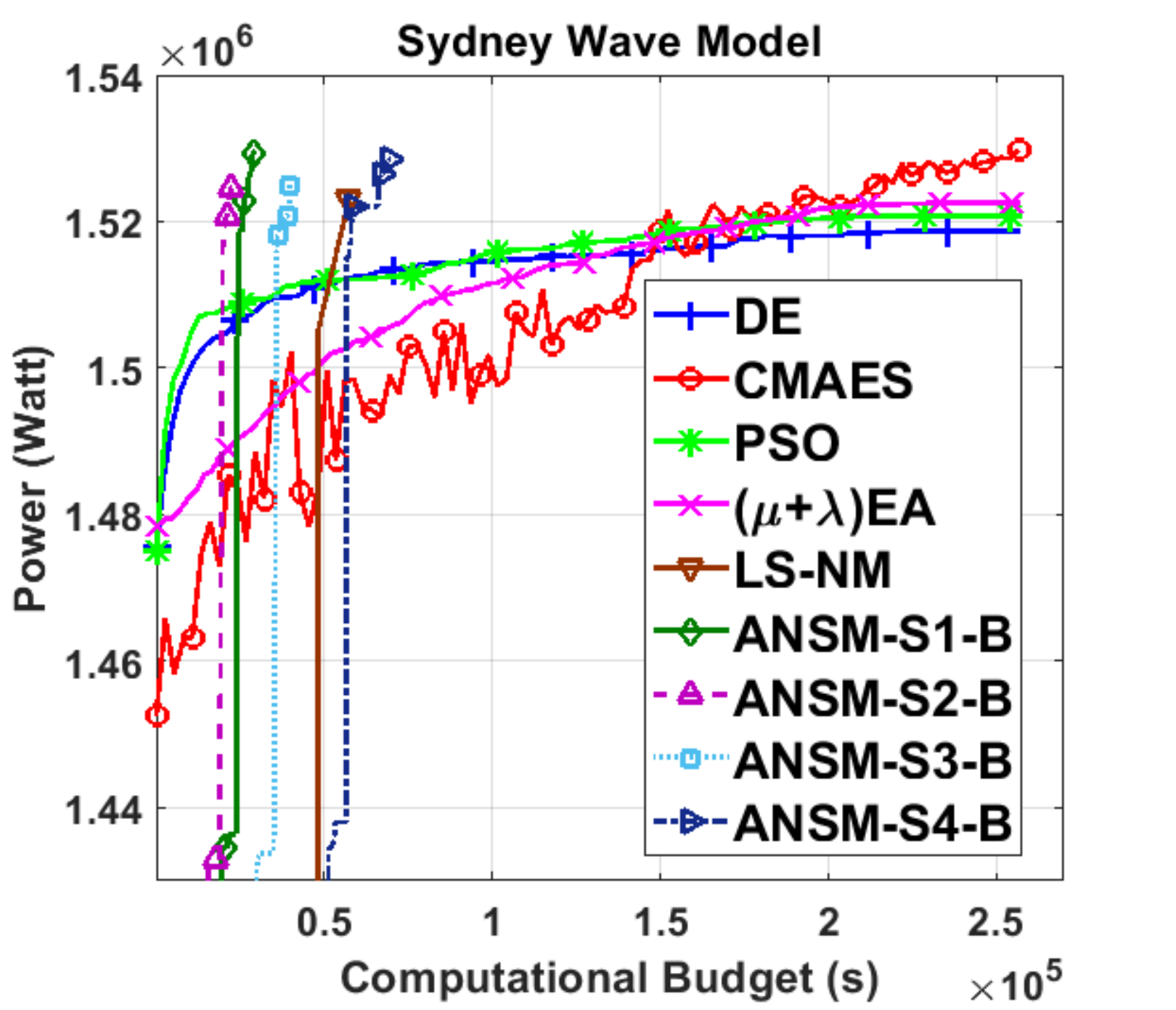}
    \includegraphics[height=7cm]{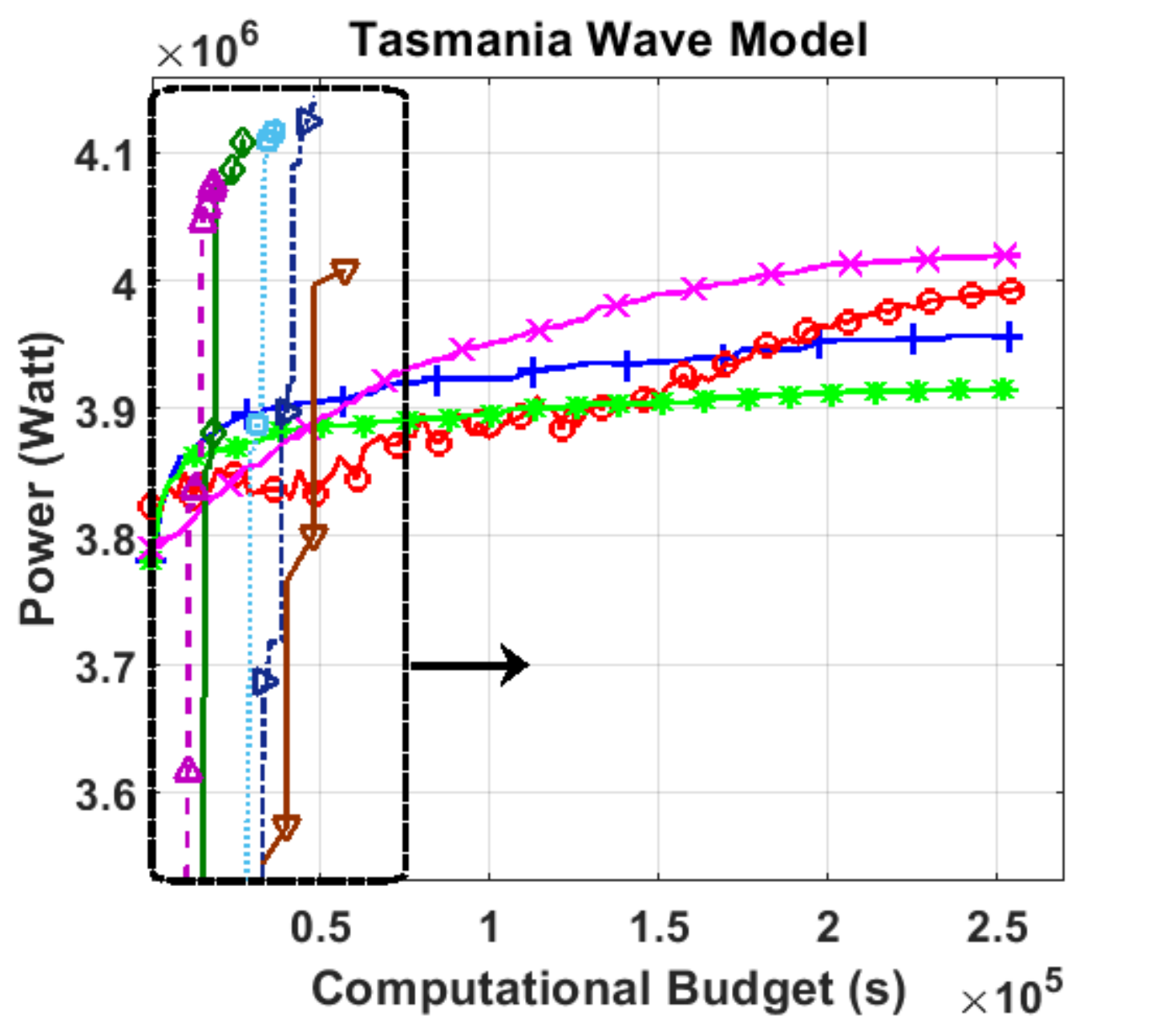}\llap{\raisebox{0.8cm}{\includegraphics[height=3.45cm]{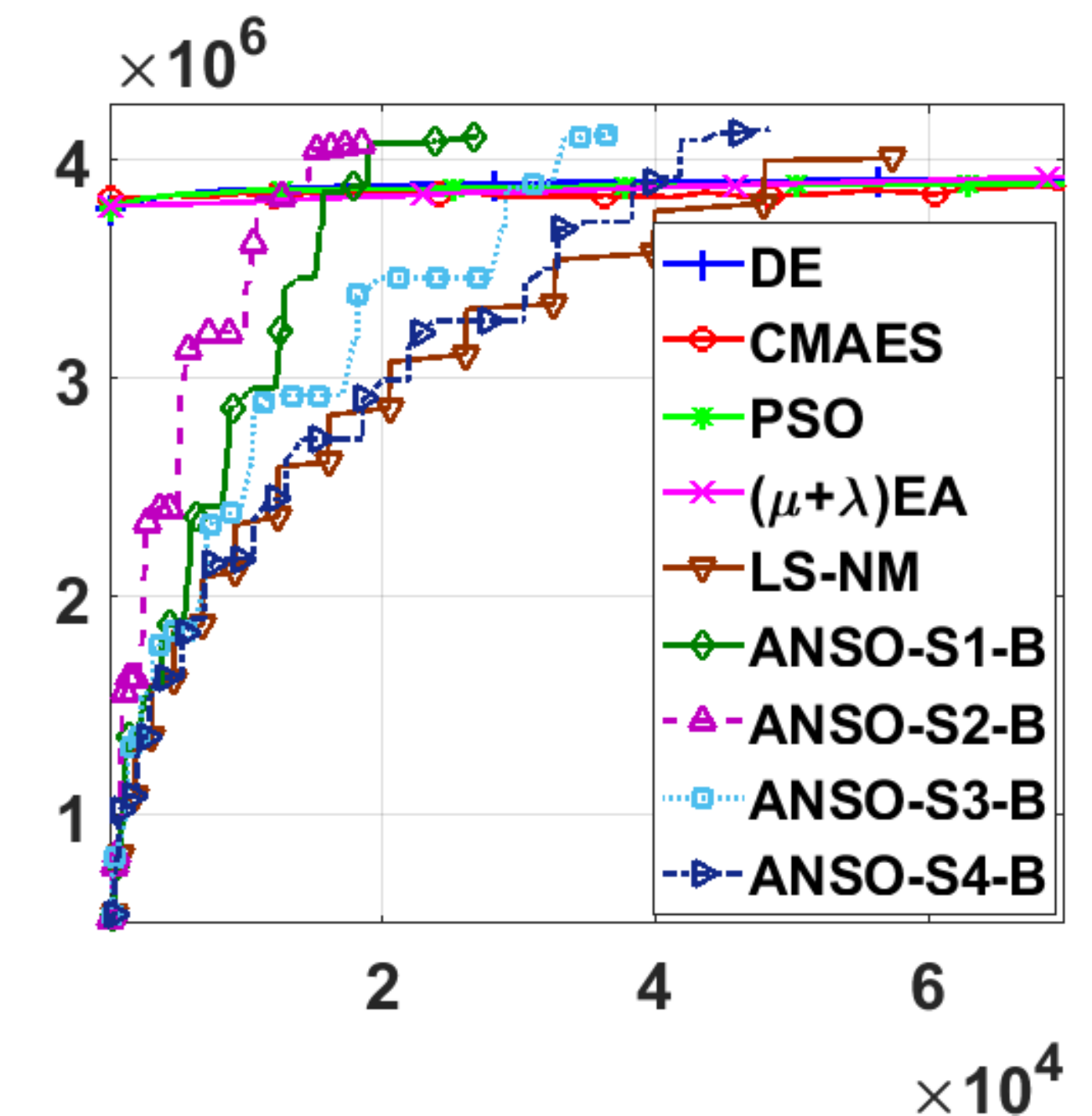}}}
  \caption{Evolution and convergence rate of the average power output contributed by the nine algorithms for four real wave models. A zoomed version of the plots is provided to show a better insight into the convergence speed.}
  \label{fig:convergence_rate}
\end{figure}
Figure~\ref{fig:convergence_rate} exhibits the convergence diagrams of the average power output of the nine compared algorithms. In all wave scenarios, ANSO-$S_2$-B has the ability to converge very fast because it estimates two sequentially placed buoys layouts power after each training process instead of evaluating one of these using the expensive simulator. ANSO-$S_2$-B is able to not only save the runtime for evaluating samples but also save the surrogate training time and is, respectively, 3, 4.5 and 14.6 times faster, on average, than ANSO-$S_4$-B, LS-NM and $(\mu+\lambda)EA$. Note again, that these timings include training and configuration times. 

\begin{figure}[t]
\centering
\includegraphics[width=\textwidth]
{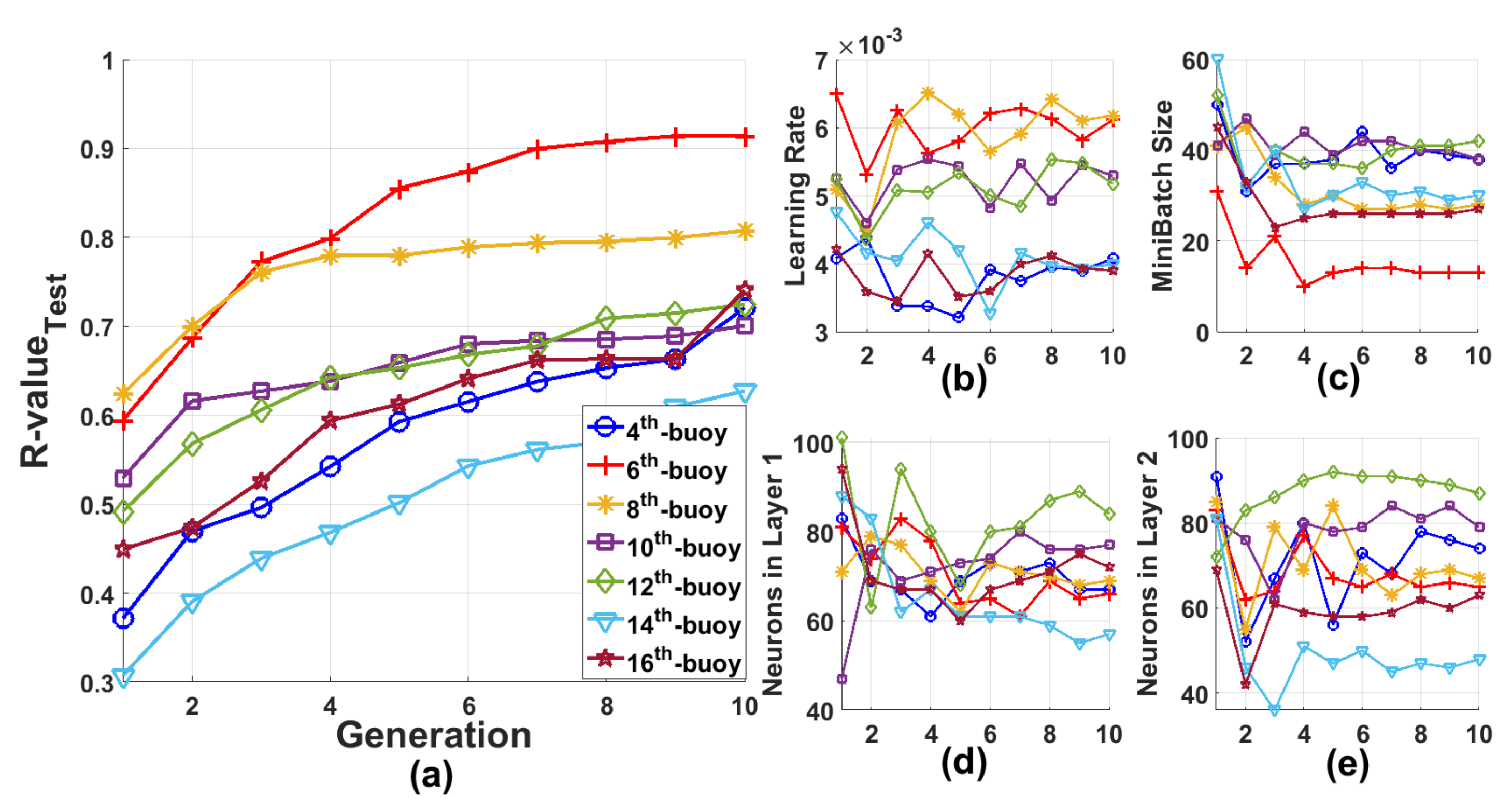}
\caption{Evolutionary (GWO) hyper-parameters optimisation: a) The vertical axis is test-set accuracy of the mean best configuration by cross-validation per generation. b) and c) show the optimisation process of the learning rate and minibatch size for estimating the power of the seven  buoys. Both d) and e) show the optimised number of neurons in the first and second LSTM layers. }\label{fig:hyper_optimization}
\end{figure}

For our neuro-surrogate model to produce accurate and reliable power estimation, we need to obtain good settings  for hyper-parameters. Finding the best configuration for these parameters in such a continuous, multi-modal and complex search space is not a trivial challenge. Figure \ref{fig:hyper_optimization} shows the GWO performance for tuning the LSTM hyper-parameters for ANSO-$S_1$.
In addition, the Pearson correlation coefficient testing values (R-value) for all trained LSTMs estimates the performance of the trained LSTM ($\bar R >=0.7$). The most challenging training process is related to the power estimation of the $14^{th}$ buoy because ANSO is faced with the boundary constraint of the search space, so the arrangement of the layouts changes. 

\section{Conclusions}

Optimising the arrangement of a large WEC farm is computationally intensive, taking days in some cases. Faster and smarter optimisation methods are needed.
We have shown that a neuro-surrogate optimisation approach, with online training and hyper-parameter optimisation, is able to outperform previous methods in terms of layout performance -- 3.2\% to 3.6\% better, respectively than $(\mu+\lambda)EA$ and LS-NM. Moreover, even including the time for training and tuning the LSTM network the neuro-surrogate model finishes optimisation faster than previous methods. Thus better results are obtained in less time -- up to 14 times faster than the $(\mu+\lambda)EA$. The approach is also highly adaptable with the model's and its hyper-parameters being tuned online for each environment. Future work will include a more detailed analysis of the training setup of the neuro-surrogate to further focus training to the sectors most relevant to buoy placement and to further adapt training for placement of buoys once the farm boundary has been reached.

\bibliographystyle{unsrt}  
\bibliography{references}

\end{document}